\documentclass[11pt]{article}
\usepackage{graphicx}
\usepackage{amssymb}
\usepackage{amscd}
\usepackage{mathrsfs}
\usepackage{longtable,lscape}
\usepackage{amsthm}
\usepackage{amsfonts}
\usepackage{amsmath}
\usepackage{bbm}
\usepackage{float}
\usepackage{url}

\newcommand{\captionfonts}{\footnotesize}
\makeatletter  
\long\def\@makecaption#1#2{%
  \vskip\abovecaptionskip
  \sbox\@tempboxa{{\captionfonts #1: #2}}%
  \ifdim \wd\@tempboxa >\hsize
    {\captionfonts #1: #2\par}
  \else
    \hbox to\hsize{\hfil\box\@tempboxa\hfil}%
  \fi
  \vskip\belowcaptionskip}
\makeatother 
\begin{document}
\title{The Quantum Nature of Identity in Human Thought: 
Bose-Einstein Statistics for Conceptual Indistinguishability}
\author{Diederik Aerts$^1$, Sandro Sozzo$^{2,1}$ and Tomas Veloz$^{3,1}$ \vspace{0.5 cm} \\ 
        \normalsize\itshape
        $^1$ Center Leo Apostel for Interdisciplinary Studies \\
        \normalsize\itshape
        and, Department of Mathematics, Brussels Free University \\ 
        \normalsize\itshape
         Krijgskundestraat 33, 1160 Brussels, Belgium \\
        \normalsize
        E-Mails: \url{diraerts@vub.ac.be, ssozzo@vub.ac.be}
          \vspace{0.5 cm} \\ 
        \normalsize\itshape
        $^2$ School of Management and IQSCS, Universit
        y of Leicester \\ 
        \normalsize\itshape
         University Road, LE1 7RH Leicester, United Kingdom \\
        \normalsize
        E-Mail: \url{ss831@le.ac.uk}
          \vspace{0.5 cm} \\ 
        \normalsize\itshape
        $^3$ Department of Mathematics, University of British Columbia, Okanagan Campus \\
        \normalsize\itshape
        3333 University Way, Kelowna, BC Canada V1V 1V7
        \\
        \normalsize\itshape
        and, Instituto de Filosof\'ia y Ciencias de la Complejidad (IFICC)  \\
        \normalsize\itshape
        Los Alerces 3024, \~Nu\~noa, Santiago, Chile  \\
        \normalsize
        E-Mail: \url{tomas.veloz@ubc.ca} \\
              }
\date{}
\maketitle
\begin{abstract}
\noindent
Increasing experimental evidence shows that humans combine concepts in a way that violates the rules of classical logic and probability theory. On the other hand, mathematical models inspired by the formalism of quantum theory are in accordance with  data on concepts and their combinations. In this paper, we investigate a novel type of concept combination were a number is combined with a noun, e.g., {\it Eleven Animals}. Our aim is to study `conceptual identity' and the effects of `indistinguishability' -- in the combination {\it Eleven Animals}, the `animals' are identical and indistinguishable -- on the mechanisms of conceptual combination. We perform experiments on human subjects and find significant evidence of deviation from the predictions of classical statistical theories, more specifically deviations with respect to Maxwell-Boltzmann statistics. This deviation is of the `same type' of the deviation of quantum mechanical from classical mechanical statistics, due to indistinguishability of microscopic quantum particles, i.e we find convincing evidence of the presence of Bose-Einstein statistics. We also present preliminary promising evidence of this phenomenon in a web-based study.
\end{abstract}

\section{Introduction\label{intro}}
There has been accumulated evidence in experimental psychology for decades indicating that the probabilistic aspects of human decision making are non-classical, in the sense that, when attempted to model them, classical probabilistic theories fail.
These experimental deviations from classical behaviour have been called `paradoxes', `fallacies', or `effects', in the literature
 \cite{tversky1982,Hampton1988a,Hampton1988b,tversky1992}.
 A new research domain called `quantum cognition' develops models of these classically unexplainable cognitive situations using the mathematical formalism of quantum mechanics~
\cite{Aerts2005a,Aerts2005b,Aerts2009a,Aerts2009b,Khrennikov2010,Aerts2011,Busemeyer2011,Busemeyer2012,Aerts2013a,Aerts2013b,Haven2013,Wang2013,Pothos2013,Aerts2013c,Aerts2014a,Sozzo2014}.  
 
We present in this paper a new, in our opinion fundamental, finding in our quantum-theoretic approach to concept theory. Our approach, originally motivated by our research on the foundations of quantum theory~\cite{Aerts1999},
led us in a first step to develop the `State Context Property' (SCoP) formalism, and use it successfully to model aspects of concepts and their combinations which classical concept theoretic approaches failed to account for. 
 In our SCoP formalism, a concept is an `entity in a specific state changing under the influence of a context', rather than a `container of instantiations', as in the traditional approaches \cite{Aerts2005a,Aerts2005b}. 
 This made possible the elaboration of a quantum-mechanical model which successfully represents different sets of data collected on conjunctions and disjunctions of two concepts \cite{Aerts2009a,Aerts2009b,Aerts2013a,Aerts2013b,Sozzo2014}. 
We show in this way that 
observed deviations from classical (fuzzy set) logic and classical probability theory can be described in terms of genuine quantum aspects such as `contextuality', `emergence', `interference', `superposition' and, more recently, entanglement \cite{Aerts2011,Aerts2013b}.

If `combining concepts' entails a dynamics reflecting the way `quantum entities are combined', one can wonder whether the analogy between `concepts as primary elements of human thought' and `quantum particles as primary elements of matter' is even deeper than what appears in 
the situations of combinations that we studied. One of the most mysterious and
not at all understood
aspects of the 
 `being of quantum entities' is `how they statistically behave when they are in situations where they are not (and cannot be, due to quantum aspects) distinguished'. Indeed, their statistical behaviour is very different from the statistical behaviour of classical objects in situations where they are not distinguished.
The latter is governed by the well-known Maxwell-Boltzmann (MB) distribution, while the former 
is described by the Bose-Einstein (BE) distribution for quantum particles with integer spin, and by the Fermi-Dirac (FD) distribution for quantum particles with semi-integer spin (we avoid considering here fractional statistical particles, for the sake of brevity). Thus, given a set of (not necessarily physical) entities, we can identify the distinguishability of such entities by observing the type of statistical distribution they obey in an experimental setting. This is exactly what we have done here for concepts, and found experimentally that BE statistics appears, thus we conclude that concepts behave as indistinguishable particles. 
Consider for example the linguistic expression ``eleven animals''. This expression 
when both ``eleven'' and ``animals'' are looked upon with respect to their conceptual structure, represents the combination of concepts {\it Eleven} and {\it Animals} into {\it Eleven Animals}, which is again a concept. 
Per definition of what a concept is, each of the {\it Eleven Animals} is completely identical on this conceptual level, and hence indistinguishable. The same linguistic expression can  however also elicit the thought about eleven objects, present in space and time, each of them then being 
an instantiation of {\it Animal}, 
and thus distinguishable from each other. Is this intuitive difference between concepts and objects the reason why the first behaves following BE statistics and the second following MB statistics? We will show further that, together with some other differences between concepts and objects, it is the case. In this sense, that we find quantum theory to be able to model concepts in a way it models identical quantum entities, is not only a strong new element for quantum cognition, but might also incorporate a new way to reflect about this mysterious and not at all understood behaviour of identical quantum entities (see, e.g.,~\cite{Castellani1998,French2006,Dieks2008}).  

In Sect. \ref{physics} we briefly summarize the statistical differences in the classical MB and quantum BE and FD distributions of physical entities. These differences are then analyzed with respect to concepts in Sect. \ref{identicalconcepts}. These sections constitute the theoretic background in which our psychological experiment in in Sect~\ref{psych} is set up. In Sect.~\ref{www} we show a preliminary study on the `World-Wide Web' that indicates a confirmation of the patterns observed in the first experiment. Each section also contains a statistical analysis of the obtained results.

\section{Identity and indistinguishability in physics\label{physics}}
We review here the basic  elements in statistical physics  which we will use in this paper for our study of identity in human concepts.

In classical mechanics, the state of an individual particle is represented by a pair $(q,p)\in \Omega$, where $q$ denotes the particle position and $p$ its momentum.  The set $\Omega$ is called the phase space of the particle. The particle evolution is then ruled by specific dynamical laws.
However, as the number of particles increases, the description of a system of these particles in terms of its individual components and states becomes more complex, and one has to introduce statistical notions (`classical statistical mechanics'). In this perspective, MB distribution predicts the probable number of particles in each of their energy states for a classical system containing many particles.

A fundamental assumption in the derivation of MB distribution is that all particles are `distinguishable', that is, one can always follow the trajectories of each particle and label them differently.  Consider, e.g., a system of distinguishable particles and suppose that these can be distributed in a given set of states of single particles. If we denote by $N$ the number of particles and by $M$ the number of single-particle states, then the total number of possible distributions is $W_{MB}(N,M)=M^{N}$. On this basis, the probability that a specific configuration $s$ is realized is $P_{MB}(s)=T_{MB}(s)/W_{MB}(N,M)$, where $T_{MB}(s)$ is the number of ways in which $s$ can be realized. One can recognize at once, in this way of calculating probabilities, that a classical Kolmogorovian framework underlies MB distribution \cite{Kolmogorov1933}. Indeed, consider a system made up of $N=2$ classical particles that can be distributed in $M=2$ states of single particle. If one applies MB distribution to this simple situation, then the number of possible arrangements is $W_{MB}(2,2)=4$, each one with a probability 1/4 of being realized. This is exactly the probabilistic situation of a system of two fair coins in a clearly Kolmogorovian framework \cite{Kolmogorov1933}.

The situation is radically different in quantum mechanics, where the state of an individual particle is represented by a unit vector $|\psi\rangle$ of a Hilbert space ${\mathscr H}$. Observables are represented by self-adjoint operators in ${\mathscr H}$, while the particle dynamics is described by `Sch{\"o}dinger's equation'. Probability occurs at a very fundamental level and does not satisfy the axioms of Kolmogorov for classical probability. This is why `quantum probability' is called `non-Kolmogorovian'. As the number of particles increases, one has to introduce a Fock space representation, and statistical considerations are needed to describe a system of many particles which are similar to those encountered in classical statistical mechanics. But, what makes quantum mechanics really different from classical mechanics are (i) the appearance of a new basic property of all microscopic entities, called `spin', (ii) the fact that 
the same quantum particles are
always `identical', or `indistinguishable' \cite{Leinaas1977,Weinberg1995}. 

That two quantum particles are identical means that one should not be able to recognize, by means of physical procedures, that an exchange has occurred between the two particles. More concretely, consider a system of two quantum particles 1 and 2 and suppose that it is represented by the unit vector $|\Psi(q_1,q_2)\rangle$ in a suitable Hilbert space. Then, indistinguishability entails that the probability of observing the system in a given region of space should be invariant when the particles are permuted or, equivalently, $|\langle\Psi(q_1,q_2)|\Psi(q_1,q_2)\rangle|^2=|\langle \Psi(q_2,q_1)|\Psi(q_2,q_1)\rangle|^2$, which entails that,  either $|\Psi(q_2,q_1)\rangle=|\Psi(q_1,q_2)\rangle$, or $|\Psi(q_2,q_1)\rangle=-|\Psi(q_1,q_2)\rangle$. In simple terms, either the wave function representing the state of the two particles remains unchanged when the particles are permuted (`symmetric wave function'), or this wave function changes sign when the particles are permuted (`anti-symmetric wave function'). A particle represented by a symmetric wave function is called `boson', a particle whose wave function is anti-symmetric is called `fermion'.

The genuinely quantum aspects of spin and indistinguishability are strictly connected by the so-called `spin-statistics theorem', which can be stated as: ``For a situation of identical integer spin particles, the wave function representing the state of such particles is symmetric. For the situation of identical half-integer spin particles, the wave function representing the state of such particles is anti-symmetric''.

It follows from the spin-statistics theorem that integer spin particles are bosons and half-integer spin particles are fermions. Moreover, the spin-statistics theorem implies that half-integer spin particles, hence fermions, are subject to the `Pauli exclusion principle', i.e. only one fermion can occupy a
specific quantum state at a specific time, which follows directly from the anti-symmetry of the wave function. For integer spin particles, i.e. bosons, with a symmetric wave function, there is instead no restriction in occupying the same state.

The above difference between fermions and bosons has a dramatic influence on the way both types of particles behave statistically. Indeed, bosons and fermions are respectively described by the `BE distribution' and the `FD distribution'. To understand the differences between these two quantum statistics and their difference with respect to MB statistics where particles are distinguishable, let us consider again the situation of $N$ particles that can be distributed in $M$ single-particle states, and suppose that the particles are identical in this case. For a system of $N$ identical bosons, the number of possible distributions is $W_{BE}(N,M)=(N+M-1)!/N!(M-1)!$, where $N!=N(N-1)(N-2)\ldots 1$. This means that fewer arrangements are available, with respect to MB case, due to indistinguishability. In the case of fermions, the Pauli exclusion principle dictates that no two fermions can be in the same state, which further reduces the number of possible distributions to $W_{FD}(N,M)=M!/N!(M-N)!$. By considering again the case $N=2=M$, we have that, for a system of 2 identical bosons, $W_{BE}(2,2)=(2+2-1)!/2!(2-1)!=3$, while the probability for each realization is 1/3. Finally, for a system of 2 identical fermions, only one realization is possible which occurs with probability 1.

The above differences between distinguishable and indistinguishable situations is statistically significant and will allow us to deduce in the next sections empirical evidences about the deviations from classicality and effects of indistinguishability of combinations of concepts.

\section{Identity and indistinguishability if concepts are considered\label{identicalconcepts}}
We analyze in this section how the notion of `indistinguishability' and `identity' can be interpreted 
for the case of concepts and how the theoretic framework 
explained in Sect. \ref{physics} can be applied. 
But, we first need to introduce the essentials of the quantum-theoretic modeling of concepts we worked out in Brussels.

With respect to applying the mathematical formalism of quantum mechanics to concepts and their dynamics, one of the major fundamental steps we took was to consider a concept as an `entity in a specific state', and not, as in the traditional approaches, as a `container of instantiations'~\cite{Aerts2005a,Aerts2005b,Aerts2009a}. The notion of `state of a concept' and the corresponding notion of `change of state' induced by context proved to be very valuable when we investigated connections between our quantum-inspired approach and traditional concept theories.

We developed a formal approach and called the `State Context Property' (SCoP) formalism~\cite{Aerts2005a,Aerts2005b}. To build a SCoP model for an arbitrary 
concept $A$ we introduce three sets, namely the set $\Sigma$ of states, the set ${\mathcal M}$ of contexts, and the set ${\mathcal L}$ of properties. The `ground state' $\hat{p}$ of the concept $A$ is the state where $A$ is not under the influence of any particular context. Whenever $A$ is under the influence of a specific context $e$, a change of the state of $A$ occurs. In case $A$ was in its ground state $\hat{p}$, the ground state changes to a state $p$. The difference between states $\hat{p}$ and $p$ is manifested, for example, by the typicality values of different exemplars of the concept, and the applicability values of properties being different in the two states $\hat{p}$ and $p$. Hence, to complete the construction of SCoP, also two functions $\mu$ and $\nu$ are introduced. The function $\mu: \Sigma \times {\mathcal M} \times \Sigma \longrightarrow [0, 1]$ is defined such that $\mu(q,e,p)$ is the probability that state $p$ of concept $A$  changes to state $q$ under the influence of context $e$. The function $\nu: \Sigma \times {\mathcal L} \longrightarrow [0, 1]$ is defined such that $\nu(p,a)$ is the weight, or normalization of applicability, of property $a$ in state $p$ of concept $A$. These mathematical structures enable the SCoP formalism to cope with both `contextual typicality' and `contextual applicability'.

What happens when we combine different concepts? 
We have studied in great detail the combination of concepts within this SCoP approach for several types of combinations (conjunctions, disjunctions, conjunctions with negations) \cite{Aerts2005a,Aerts2005b,Aerts2009a,Aerts2011,Aerts2013a,Aerts2013b,Sozzo2014}. However, if the conceptual expression combines a number, e.g. {\it Eleven} with a noun,  e.g. {\it Animals}, a completely new `structure of combining' comes into being, which, as we will analyse now, is provoked by the aspect of `identity for concepts'. Even in the realm of a pure theoretical analysis, which is what we will do here now, the procedure of combining a number with a noun precludes the presence of a BE statistics rather than a MB statistics. The experimental results that we present in sections \ref{psych} and \ref{www} contain evidence for this and will also be analysed accordingly. 
${\cal L}$, and $\nu$, in the structures of SCoP do not play a role in what follows, hence we can concentrate on $\Sigma$, ${\cal M}$ and $\mu$. We keep along our example of {\it Eleven Animals} to explain this new combination structure. Suppose we consider two states of {\it Animal}, namely {\it Cat} and {\it Dog}, and hence the situation where {\it Eleven Animals} can be {\it Cats} or {\it Dogs}, and we make now a strict exercise to `reason on the level of the concepts' and not on the level of their `instantiations into objects'. Then, the conceptual meaning of {\it Eleven Animals}, which can be {\it Cats} or {\it Dogs} gives rise in a unique way to twelve possible states, if the criterion is that this meaning is realised in each state. Lets denote them by $p_{11,0}$, $p_{10,1}$, \ldots, $p_{1,10}$ and $p_{0,11}$, and they stand respectively for {\it Eleven Cats} (and no dogs), {\it Ten Cats And One Dog}, \ldots, {\it One Cat And Ten Dogs} and {\it Eleven Dogs} 
(and no cats). 
We want to investigate the `probabilities of change of the ground state $\hat p$ of the combined concept {\it Eleven Animals} into one of the twelve states $p_{11,0}$, $p_{10,1}$, \ldots, $p_{1,10}$ and $p_{0,11}$' in two specific cases, 
tested both experimentally. The first case is where human subjects are asked, after they have been presented the twelve states as the set of situations they can choose from, which state is their preferred one. The relative frequency arising from their answers (Sect. \ref{psych}) we consider as the probabilities of change of the ground state $\hat p$, to the chosen state, which is one of the set of states in $\{p_{11,0}$, $p_{10,1}$, \ldots, $p_{1,10}$, $p_{0,11}\}$. The context $e$ involved in this experiment is 
difficult to characterize 
in great detail~\cite{Aerts2005a}, but important is to notice that it is mainly determined by the  
`combination procedure of the concepts {\it Eleven} and {\it Animals}'  
and the `meaning contained in the new combination'  
for participants in the experiment. Hence, our psychological experiment tests whether participants follow the `conceptual meaning' of {\it Eleven Animals} treating {\it Dogs} as identical, and also {\it Cats} as identical, or participants follow the `instantiations into objects meaning' of {\it Eleven Animals} treating {\it Dogs} as distinguishable, and also {\it Cats} as distinguishable.
The second case consists of a preliminary exploration of the frequency of conceptual elicitations in the World-Wide Web through the automatic use of a search engine. Namely, for each state in $\{p_{11,0}$, $p_{10,1}$, \ldots, $p_{1,10}$, $p_{0,11}\}$, we select linguistic expressions that most commonly refer to the state, and use the search engine to estimate the total number of web-documents where these sentences are elicited. The relative frequency of the number of web-documents obtained for each state we consider as the probability of changing from the ground state $\hat p$, to the chosen state. The context $e$ that provokes the change of state in this case is again difficult to characterize in detail, but is defined by the entirety of elements that made someone write the passage of text where the combination was encountered.  Indeed, the sentences we consider in our experiment were chosen so they indicate the most typical use of the concept combination.  This means that our computational experiment tests also whether the elicitation of concept combinations in web-documents tends to follow the `conceptual meaning', or the `instantiations into objects meaning' of {\it Eleven Animals}.

Mathematically, we are thus representing the conceptual entity {\it Eleven Animals} by the SCoP model $(\Sigma,{\cal M},
\mu
)$, where 
$\Sigma=\{\hat p, p_{11,0}, p_{10,1}, \ldots, p_{1,10}, p_{0,11}\}$, ${\cal M}= \{ e \}$, and the considered transition probabilities are $\{\mu(q,e,\hat p)\ \vert q\in \{p_{11,0}, p_{10,1}, \ldots, p_{1,10}, p_{0,11}\}\}$. We recognise right away the situation encountered in Sect. \ref{physics} in the structure of $\mu(q,e,\hat p)$. Indeed, this conceptual situation is analogous to the one in which one has $N=11$ particles that can be distributed in $M=2$ possible states. It is 
hence possible, 
by just looking at the relative frequencies obtained in the experiment, to find out whether a classical MB statistics or a quantum-type, i.e. BE statistics, applies to this situation.

Let us calculate concretely what both MB and BE give in the specific case of {\it Eleven Animals}. In case that MB would apply, it would mean that things happen as if there are underlying the twelve states hidden possibilities, namely $T(n,C;11-n,D)=11!/n!(11-n)!$ in number, for the specific state of $n$ {\it Cats} and $11-n$ {\it Dogs}, $n=0, \ldots,11$. Of course, this ``is'' true in case the cats and dogs are real cats and dogs, hence are `objects existing in space and time', which is why for objects in the classical world indeed MB statistics applies. Let us calculate the probabilities involved then. 
For the sake of simplicity, we assume the existence of two probability values $P_{Cat}$ and $P_{Dog}$, such that $P_{Cat}+P_{Dog}=1$, and that the events of 
making actual such an underlying state for {\it Cat} and {\it Dog} are independent. Hence the probability 
for $n$ exemplars of {\it Cat} and $11-n$ exemplars of {\it Dog} is
then $\mu_{MB}^{P_{Cat},P_{Dog}}(p_{n,11-n},e,\hat p)=T(n,C;11-n,D) P_{Cat}^nP_{Dog}^{11-n}$. Note that, under 
the assumption of MB statistics, $\mu_{MB}^{P_{Cat},P_{Dog}}(p,e,\hat p)$ becomes the binomial probability distribution. For example, if $P_{Dog}=P_{Cat}=0.5$, the number of possible arrangements for the state {\it Eleven Cats And Zero Dogs} and for the state {\it Zero Cats And Eleven Dogs} is 1, hence the corresponding probability for these configurations is $\mu_{MB}^{P_{Cat},P_{Dog}}(p_{0,11},e,\hat p)=\mu_{MB}^{P_{Cat},P_{Dog}}(p_{11,0},e,\hat p)=0.0005$. Analogously, the number of possible arrangements for the state {\it Ten Cats And One Dog} and for the state {\it One Cat And Ten Dogs} is 11, hence the corresponding probability for these configurations is $\mu_{MB}^{P_{Cat},P_{Dog}}(p_{10,1},e,\hat p)=\mu_{MB}^{P_{Cat},P_{Dog}}(p_{1,10},e,\hat p)=0.0054$, and so on. When $P_{Cat}$ and  $P_{Dog}$ are  equal, MB distribution entails a maximum value for such a probability. In this example, this corresponds to the situation of {\it Six Cats And Five Dogs} and {\it Five Cats And Six Dogs} with $\mu_{MB}^{P_{Cat},P_{Dog}}(p_{6,5},e,\hat p)=\mu_{MB}^{P_{Cat},P_{Dog}}(p_{6,5},e,\hat p)=0.2256$. 
Suppose that we would visit farms, all of them guarding at least 11 cats and 11 dogs, and carry with us  12 cages. And 
then we would pick with probabilities respectively $P_{Cat}$ and $P_{Dog}$ a cat or a dog and in this way fill  each of the cages with a combination of 11 cats and dogs. We would then find exactly the above calculated MB statistics.

Let us now make the calculation for BE statistics, where we keep making the exercise of only reasoning on the level of concepts, and not on the level of instantiations. 
Then, for the twelve different states do not exist underlying hidden states, because the existence of such states would mean that we reason on more concrete forms in the direction of instantiations 
-- we, for example, forbid in this case to consider the features `white' and `black' to apply for the concept {\it Animal} because this will allow to eventually distinguish the cats for state $p_{1,2}=${\it Two Cats and Nine Dogs}. Also in this case, suppose that {\it Cat} and {\it Dog} have an independent elicitation probability  $P_{Cat}$ and $P_{Dog}$, thus $P_{Cat}+P_{Dog}=1$. Thus, the probability that there are $n$ exemplars of {\it  Cat} and $(11-n)$ exemplars of {\it Dog} is $\mu_{BE}^{P_{Cat},P_{Dog}}(p_{n,11-n},e,\hat p)=  (nP_{cat}+(11-n)P_{dog})/(\frac{12\times 11}{2})$. Note that as $P_{Cat}=1-P_{Dog}$, then $\mu_{BE}^{P_{Cat},P_{Dog}}(p_{n,11-n},e,\hat p)$ is a linear function. Moreover, when 
$P_{Cat}=P_{Dog}=0.5$, we have that $\mu_{BE}(p_{n,11-n},e,\hat p)=1/12$ for all values of $n$, recovering BE distribution explained in Sect. \ref{physics}.

The above analysis shows that, if one performs experiments on a collection of concepts like {\it Eleven Animals} to estimate the probability of elicitation for each state, then 
it is possible to establish whether a distribution of MB-type $\mu_{MB}^{P_{Cat},P_{Dog}}(p_{n,11-n},e,\hat p)$, or of BE-type $\mu_{BE}^{P_{Cat},P_{Dog}}(p_{n,11-n},e,\hat p)$, or a very different one, holds. However, in case there are strong deviations from a MB statistics, while a quasi-linear distribution is obtained, then this would indicate that, in context $e$, where only {\it Cat} and {\it Dog} are allowed to be states of the concept {\it Animal}, the statistical distribution of the collection of concepts {\it Eleven Animals} is of a BE-type and that concepts present a quantum-type indistinguishability.

The above analysis also provides us with a view of what can possibly take place and what is being tested in the experiments, and even makes it possible to put forward some element of prediction with respect to the outcomes of the experiments. The crucial question is `how and in which amount the human mind treats the cats and dogs as identical in the way our theoretical analysis indicates'. More concretely, if a human mind `imagines' a combination of, lets say 2 cats and 9 dogs, and another combination of 5 cats and 6 dogs. Does the human mind then takes into account that `if it is about real cats and dogs', there are many more ways to put 5 cats and 6 dogs into  a cage, than there are ways to put 2 cats and 9 dogs into  a cage, when this human mind is asked to elect one of these two combinations according to preference.

We believe that the answer to the above question is `no', and that this is the reason that BE statistics appears and not MB statistic for the situation of identical concepts. But, let us be more specific, and at the same time derive this prediction for our experiments. Suppose it is not about cats and dogs, but about sisters and brothers. Then the human mind asked to do the test of electing between the two cases mentioned above might be inclined to be influenced by real families that he or she knows, with 11 children, and  then of course, like we also explained above, in the real world there are much more situations of families with 5 brothers an six sisters, than there are with 2 brothers and 9 sisters. So, our prediction with respect to the outcome of the experiments is that `the more the concepts and states of them are easily in imagination connected to real life situations, the more MB statistics might show up on top of an underlying BE statistics, which is the one that would appear in pure form if the human mind is not influenced at all in its imagining of the different states (combinations) by comparing to real life existing situations'.
Our considerations in the last two sections provide the theoretical background where the experiments we performed are set. We will devote Sects. \ref{psych} and \ref{www} to introduce, describe and analyze these experiments.

\section{Psychological evidence of indistinguishability in concepts\label{psych}}
The experiment involved 88 participants. We considered a list of concepts 
$A^i$ of different (physical and non-physical) nature, $i=1, \ldots,14$, and two possible exemplars (states) $p_{1}^i$ and $p_{2}^i$ for each concept. Next we requested participants to choose one exemplar of a combination 
$N^{i} A^i$ of concepts, where 
$N^{i}$ is a natural number. The exemplars of these combinations of concepts 
$A^i$ are the states
$p^i_{k,N^{i}-k}$ describing the conceptual combination   `$k$ exemplars in state $p^i_1$  and $(N^i-k)$ exemplars in state $p^i_2$', where $k$ is an integer such that 
$k=0, \ldots, N^i$. For example, the first collection of concepts we considered is  $N^1 A^1$ corresponding to the compound conceptual entity {\it Eleven Animals}, with $p^i_{1}$ and $p^i_{2}$ describing the exemplars  {\it Cat} and {\it Dog} of the individual concept {\it Animal}, respectively, and  $N^1=11$. The exemplars (states) we considered are thus $p^1_{11,0}$, $p^1_{10,1}$, \ldots, $p^{1}_{1,10}$, and $p^{1}_{0,11}$, describing the combination  {\it Eleven Cats And Zero Dogs}, {\it Ten Cats And One Dog}, \ldots, {\it One Cat And Ten Dogs}, and  {\it  Zero Cats And Eleven Dogs}. The other collections of concepts we considered in our psychological experiment are reported in Table 1.
\begin{table}[H]\label{categories-psych}
\begin{center} 
\begin{tabular}{|c|c|c|c|c|} \hline
$i$ & $N^i$ & $A^i$ & $p_{1}^i$ & $p_{2}^i$  \\ 
\hline 
\hline
1& 11 & {\it Animals} & {\it Cat} & {\it Dog} \\ \hline
2& 9 & {\it Humans} & {\it Man} & {\it Woman}\\ \hline
3& 8 & {\it Expressions of Emotion} & {\it Laugh} & {\it Cry} \\ \hline
4& 7& {\it Expressions of Affection} & {\it Kiss} & {\it Hug}\\ \hline
5& 11& {\it Moods} & {\it Happy} & {\it Sad} \\ \hline
6& 8 &{\it Parts of Face} & {\it Nose} & {\it Chin} \\ \hline 
7& 9 & {\it Movements} & {\it Step} & {\it Run} \\ \hline
8& 11 & {\it Animals} & {\it Whale} & {\it Condor}\\ \hline
9& 9 & {\it Humans} & {\it Child} & {\it Elder} \\ \hline
10& 8 & {\it Expressions of Emotion} & {\it Sigh} & {\it Moan} \\ \hline
11& 7& {\it Expressions of Affection} & {\it Caress} & {\it Present}\\ \hline
12& 11& {\it Moods} & {\it Thoughtful} & {\it Bored}\\ \hline
13& 8 &{\it Parts of Face} &{\it Eye} & {\it Cheek} \\ \hline
14& 9 &{\it Movements} & {\it Jump} & {\it Crawl} \\ \hline
\end{tabular}
\caption{List of concepts and their respective states for the psychological concept on identity and indistinguishability.}
\end{center} 
\end{table}

We computed the parameters $P^{MB}_{p_{1}^i}$ and $P^{BE}_{p_{1}^i}$ that minimize the the R-squared value of the fit using the  distributions $\mu_{MB}^{P_{p_{1}^i},P_{p_{2}^i}}$ and $\mu_{BE}^{P_{p_{1}^i},P_{p_{2}^i}}$ for each $i=1, \ldots, 14$. Hence, we fitted the distributions obtained in the psychological experiments using MB and BE statistics presented in Sect. \ref{physics}
(note that only one 
parameter is needed as $P^{MB}_{p_{2}^i}=1-P^{MB}_{p_{1}^i}$ and $P^{BE}_{p_{2}^i}=1-P^{BE}_{p_{1}^i}$).
Next, we used the `Bayesian Information Criterion (BIC)'  \cite{KASSBIC} to estimate which model provides the best fit and contrast this criterion with the R-squared value.
 Table
2 summarizes the statistical analysis. The first column of the table identifies the concept in question (see Table 1),
the second and third columns show $P^{MB}_{p_{1}^i}$ and the $R^2$ value of the MB statistical fit, the fourth and fifth columns show $P^{BE}_{p_{1}^i}$ and the $R^2$ value of the BE statistical fit. The sixth column shows the $\Delta_{\textrm{BIC}}$ criterion to discern between the $\mu_{MB}^{P_{p_{1}^i}P_{p_{2}^i}}$ and $\mu_{BE}^{P_{p_{1}^i}P_{p_{2}^i}}$, and the seventh column identifies the distribution which best fits the data for concept  $A^i$, $i=1, \ldots,14$.
\begin{table}[H]
\begin{center} 
\begin{tabular}{|c|c|c|c|c|c|c|c|} \hline
$i$&$P^{MB}_{p_{1}^i}$&$R_{MB}^2$&$P^{BE}_{p_{1}^i}$&$R_{BE}^2$&$\Delta_{\textrm{BIC}}$ &Best Model  \\ \hline \hline
1& 0.55 &-0.05 &0.16 &0.78 &19.31  & BE strong \\ \hline
2& 0.57 &{\bf 0.78} &0.42 &0.44 &-9.54 &MB strong \\ \hline
3& 0.82 &0.29 &{\bf 0.96} &0.79 & 10.81&BE strong \\ \hline
4& 0.71 &0.81 &0.53 &0.77 &-1.69 & MB weak\\ \hline
5& 0.25 &{\bf 0.79} &0.39 &0.93 & 14.27&BE strong\\ \hline
6& 0.62 &0.59 &0.61 &0.57 &-0.37 &MB weak\\ \hline
7& 0.72 &0.41 &0.64 &{\bf 0.83} &12.66 &BE strong\\ \hline
8& 0.63 &0.58 &0.47 &0.73 &5.53 & BE positive\\ \hline
9& 0.45 &{\bf 0.87} &0.26 &0.67 &-9.69 &MB strong\\ \hline
10&0.59 &0.50 &0.63 &0.77 &7.17 &BE positive\\ \hline
11&0.86 &0.46 &1.00 &{\bf 0.87} &11.4 &BE strong \\ \hline
12&0.21 &0.77 &0.00 &0.87 &6.68 &BE positive\\ \hline
13&0.62 &0.54 &0.71 &0.67 &2.97 &BE weak \\ \hline
14&0.81 &0.20 &0.91 &{\bf 0.90} &20.68 &BE strong\\ \hline
\end{tabular}
\label{stat-results}
\caption{Results of statistical fit for the psychological experiment. Each column refers to the 14 collections of concepts introduced in Table 1.}
\end{center} 
\end{table}
Note that, according to the BIC criteria, negative $\Delta_{\textrm{BIC}}$ values imply that the category is best fitted by a MB distribution, whereas positive $\Delta_{\textrm{BIC}}$ values on row $i$ imply the concept 
$A^i$ is best fitted with a BE distribution. Moreover, when $|\Delta_{\textrm{BIC}}|<2$ there is no clear difference between the models, when $2<|\Delta_{\textrm{BIC}}|<6$ we can establish a positive but not strong difference towards the model with smallest value, whereas when $6<|\Delta_{\textrm{BIC}}|$ we are in presence of strong evidence that one of the models provides better fit than the other model~\cite{KASSBIC}. We see that categories $2$ and $9$ show a strong $\Delta_{\textrm{BIC}}$ value towards MB-type of statistics, and that categories $1,3,5,7,11,12$ and $14$ show a strong $\Delta_{\textrm{BIC}}$ value towards BE-type of statistics.
Complementary to the BIC criterion, the $R^2$ fit indicator helps to see whether or not the indications of $\Delta_{\textrm{BIC}}$ can be confirmed with a good fit of the data. Interestingly, the concepts we have identified with strong indication towards one type of statistics have $R^2$ values larger than $0.78$ (such $R^2$ values are marked in bold text), which indicates a fairly good approximation for the data. Moreover, note that in all the cases with strong tendency towards one type of statistics, the $R^2$ of the other type of statistics shows is poor. This confirms the fact that we can discern between the two types of statistics depending on the concept in question.
The results anticipated in Sect. \ref{intro} are thus confirmed by our statistical analysis in this section. 
Conceptual combinations exists, like {\it Nine Humans}, whose distribution follows MB statistics. However, also conceptual combinations, like {\it Eleven Animals}, {\it Eight Expressions of Emotion} or {\it Eleven Moods}, whose distribution follows BE statistics exist. The conclusion is that the nature of identity in these concept combinations is of a quantum-type and 
in these combinations the human mind treats the two states we consider as identical and indistinguishable.  Also the hypothesis that `the more easy the human mind 
imagines spontaneously instantiations, e.g. {\it Nine Humans}, the more MB, and the less easy such instantiations are activated in imagination, e.g. {\it Eight Expressions of Emotion}, the more BE statistics appears' is confirmed by our experiment.

\section{Preliminary study of indistinguishability in concepts on the web\label{www}}
The results found on the psychological experiment suggest that indistinguishability is a default manner to elicit collections of concepts. We investigated if this phenomena can also be found on the web.  Namely, we chose $N^{i}$  to be a number between $2$ and $15$, and four pairs of states
 $(p_{1}^j$, $p_{2}^j)$ for each $j=1, \ldots, 4$, each pair having states of a common concept. For example, the states $p_{1}^1$ and $p_{2}^1$ correspond to {\it Cat} and {\it Dog}, that are commonly understood to be states
of the concept {\it Animal}. Next, for each number $3 \leq k \leq N^{i}$ and pair $(p_{1}^{j},p_{2}^{j})$ of states, we built a set of sentences $r^{j}_{k,N^{i}-k}$ that refer to the state $p_{k,N^{i}-k}^{j}$ similar to the psychological experiment. For example, the state $p_{1,3}^{1}$ describing {\it Three Cats And One Dog} is referred by the sentences  $r^{1}_{1,3}=\{$``three cats and one dog'', ``one dog and three cats''$\}$. The language entities chosen for this experiment are shown in Table 3, and the list of references in Table 4.
\begin{table}[H]
\begin{center} 
\begin{tabular}{|c|c|c|} 
\hline
$j$& $p_{1}^{j}$&$p_{2}^{j}$ \\
 \hline
1 & ``cat''/``cats'' & ``dog''/``dogs''\\ \hline
2 & ``man''/``men'' &  ``woman''/``women''\\ \hline
3 & ``win''/``wins'' & ``loss''/``losses''\\ \hline
4 & ``son''/``sons'' & ``daughter''/``daughters'' \\ \hline
\end{tabular}
\label{Concepts-states-web}
\caption{List of singular/plural reference to states used to perform the web-based experiment.}
\end{center} 
\end{table}
 The number of webpages where each sentence $r^{j}_{k,N^{i}-k}$ appears was computed with the Bing search API for web developers.\footnote{For more information, see \url{http://datamarket.azure.com/dataset/bing/search}.}

This study can only be considered preliminary, due to the small sample size and to well-known difficulties of computational semantics (for a complete review of these difficulties see~\cite{Pylyshyn1984}). In our SCoP setting, some of these difficulties involve:\footnote{The possibility of the study of meaning using computational techniques in the context of SCoP  is mentioned   in~\cite{Aerts2013b}, and  planned to be elaborated in great detail in~\cite{tveloz-CL}.}

(a)  A state can be referred by potentially infinite linguistic expressions.

(b) A state can be linked to potentially infinite concepts.

(c) Context in a conceptual reference is hard to trace.

Although these limitations are strong, we believe that our preliminary analysis  of the issue of distinguishability of concepts on the web contains already strong evidence for BE rather than MB.

We summarize the results of our computational experiment in Table 5. The first column specifies $N^{i}$, the other four columns specify the pair of states, according to Table 3, used in the experiment. Each element of the table contains a pair of numbers. The first number correspond to  $\Delta_{\textrm{BIC}}$, and the second number to the $R^2$ of the best fit according to the $\Delta_{\textrm{BIC}}$ criteria, i.e. when $\Delta_{\textrm{BIC}}<0$, then the $R^2$ value is the one from MB statistics, and when a positive $\Delta_{\textrm{BIC}}$ is shown, then the second number of the pair is the $R^2$ value for the BE statistics. Moreover, if $R^2<0.65$ we show instead a character $-$ to represent the fact that $R^2<0.65$ cannot be considered a significant fit. The last row summarizes the behaviour of the pair of states. 
\begin{table}[H]
\begin{center} 
\begin{tabular}{|c|c|} \hline
$N^{i}$& List of references\\ \hline
0& ``0'',``no'',``zero''\\ \hline
1& ``1'',``a'',``one''\\ \hline
2& ``2'',``two'',``a couple of''\\ \hline
3& ``3'',``three''\\ \hline
4& ``4'',``four''\\ \hline
5& ``5'',``five''\\ \hline
6& ``6'',``six''\\ \hline
7& ``7'',``seven''\\ \hline
8& ``8'',``eight''\\ \hline
9& ``9'',``nine''\\ \hline
10& ``10'',``ten''\\ \hline
11& ``11'',``eleven''\\ \hline
12& ``12'',``twelve''\\ \hline
13& ``13'',``thirteen''\\ \hline
14& ``14'',``fourteen''\\ \hline
15& ``15'',``fifteen''\\ \hline
16& ``16'',``sixteen''\\ \hline
\end{tabular}
\label{number-reference-web}
\caption{List of references to numbers used to perform web-based experiment.}
\end{center} 
\end{table}
\begin{table}[H]
\begin{center} 
\begin{tabular}{|c|c|c|c|c|} \hline
$N^i$&$j=1$&$j=2$&$j=3$ &$j=4$  \\ \hline \hline
3& $-3.9,0.79$&$-4.4,0.82$&$-1.5,-$&$-3.9,0.80$ \\ \hline
4&$-8.9,0.92$&$-7.4,0.91$&$-4.2,0.80$&$-9.2,0.93$  \\ \hline
5& $-10.5,0.94$&$-3.83,0.81$&$-8.20,0.90$&$-14.7,0.97$ \\ \hline
6&$-5.0,0.84$&$-3.6,0.82$&$-2.6,0.80$&$-15.5,0.96$  \\ \hline
7&$-2.0,0.77$&$3.1,0.72$&$-1.5,0.75$&$-4.7,0.85$  \\ \hline
8&$2.0,0.72$&$-0.1,0.74$&$-0.8,0.77$&$-1.5,0.79$  \\ \hline
9&$5.5,0.69$&$7.3,0.76$&$6.4,0.78$&$-7.4,0.87$  \\ \hline
10&$9.0,0.70$&$0.5,-$&$10.5,0.77$&$-11.4,0.89$  \\ \hline
11&$2.4,-$&$10.0,-$&$9.3,0.73$&$-5.2,0.80$  \\ \hline
12&$10.4,0.70$&$7.0,0.72$&$11.1,0.72$&$-6.4,0.79$  \\ \hline
13&$6.6,-$&$11.1,-$&$12.7,0.76$&$9.4,-$  \\ \hline
14&$13.6,-$&$17.3,0.71$&$10.8,0.76$&$-8.9,-$  \\ \hline
15&$9.1,-$&$23.0,0.79$&$2.3,-$&$-17.6,0.8$\\ \hline
Type& Mixed&Mixed&Mixed&MB only\\ \hline
\end{tabular}
\label{stat-results-web}
\caption{Results of statistical fit of web-based experiment.}
\end{center} 
\end{table}
We can identify three concrete trends:

(i) When $3\leq N^i\leq 6$ the four pairs of states are of MB-type.

(ii) When $7\leq N^i\leq 10$, three pair of states `shift' to BE-type.

(iii) For $N^i\in\{11,13,14,15\}$, at least two pairs of states show a poor $R^2$ fit.

The first two trends indicate that for $N^i \leq 6$, the concepts in the combination are treated as distinguishable entities, while for $7\leq N^i$ concepts tend to `shift' towards the BE statistics. Hence we conclude that, when numbers are large enough, humans tend to treat collections of concepts as indistinguishable entities. This is consistent with the fact that we cannot propely process (e.g. remember, repeat, compare) intermediate or large collections of distinguishable entities. However, by not trying to distinguish the entities when elicited in large collections, we make use of language properly communicate and cope with them.  More specifically we use the potential of language to express concepts as identical and indistinguishable, while, without loss of meaning, transfer genuinely what is meant to be communicated. The third trend shows that for some numbers above ten, the MB and BE distributions deviate significantly from the data. We believe this effect is due to the infrequent appearance of certain configutations of states on the web.  

\section{Conclusions}
We investigated the type of statistics underlying a conceptual combination of a number and a noun. By performing psychological and web experiments we found that, in several cases, quantum statistics, and in particular BE-type statistics, provides a much better estimation of the data when compared to the classical MB statistics. Hence, we conclude that, when a number-concept is combined with a noun-concept, expressing an amount indicated by this number of exemplars of the noun-concept, these exemplars are considered to be identical and indistinguishable. This result supports the `quantum cognition hypothesis' stating that the structure and dynamics of conceptual entities strongly resembles the structure and dynamics of quantum particles. On one hand, our results invite to deepen investigation on conceptual distinguishability using the SCoP theory of concepts. Particularly regarding how to perform experiments where i) concepts can only be elicited in real space-time situations, and hence an influence of MB-type statistics appears, and ii) concepts can only be elicited in a purely semantic realm, and as a consequence a BE-type of statistics is dominant. We believe that the core of the  identity and indistinguishability issue for concepts lies here. On the other hand, our results invite quantum  theory researchers to 
reflect about the issue of  identity and indistinguishability for quantum particles taking into account the way identity and indistinguishability appear in the conceptual realm \cite{Aerts2009b}.

\end{document}